\documentclass[10pt,twocolumn,letterpaper]{article}

\usepackage{cvpr}              
\usepackage{amsmath}
\usepackage{amssymb}
\usepackage{pifont}
\usepackage{multirow}
\usepackage{multicol}
\usepackage{graphicx}
\usepackage{makecell}
\usepackage{color}
\usepackage{float}
\usepackage{colortbl}
\usepackage[accsupp]{axessibility}
\usepackage{xcolor}
\usepackage{bm}
\graphicspath{{images/}}

\usepackage{ algorithmic}
\usepackage[linesnumbered,boxed,ruled,commentsnumbered]{algorithm2e}

\PassOptionsToPackage{prologue,dvipsnames}{xcolor}
\usepackage[dvipsnames]{xcolor}

\definecolor{cvprblue}{rgb}{0.21,0.49,0.74}
\usepackage[pagebackref,breaklinks,colorlinks,citecolor=cvprblue]{hyperref}

\def\paperID{3748} 
\def\confName{CVPR}
\def\confYear{2024}

\title{Decoupled Pseudo-labeling for Semi-Supervised Monocular 3D Object Detection}

\author{Jiacheng Zhang$^{1}$\thanks{Equally-contributed authors. Work done during an internship at Baidu.} \quad
Jiaming Li$^{1*}$ \quad
Xiangru Lin$^{2*}$ \quad
Wei Zhang$^{2}$ \quad \\
Xiao Tan$^{2}$ \quad
Junyu Han$^{2}$ \quad Errui Ding$^{2}$ \quad Jingdong Wang$^{2}$ \quad
Guanbin Li$^{1,3}$\thanks{Corresponding author.} \\
$^{1}$School of Computer Science and Engineering, Sun Yat-sen University, Guangzhou, China\\
$^{2}$Department of Computer Vision Technology (VIS), Baidu Inc., China\\
$^{3}$GuangDong Province Key Laboratory of Information Security Technology\\
{\tt\small \{zhangjch58,lijm48\}@mail2.sysu.edu.cn, \{liguanbin\}@mail.sysu.edu.cn } \\
{\tt\small \{linxiangru,zhangwei99,tanxiao01,hanjunyu,dingerrui,wangjingdong\}@baidu.com}
}

\begin{document}
\maketitle
\begin{abstract}
 We delve into pseudo-labeling for semi-supervised monocular 3D object detection ~(SSM3OD) and discover two primary issues: a misalignment between the prediction quality of 3D and 2D attributes and the tendency of depth supervision derived from pseudo-labels to be noisy, leading to significant optimization conflicts with other reliable forms of supervision. To tackle these issues, we introduce a novel decoupled pseudo-labeling ~(DPL) approach for SSM3OD. Our approach features a Decoupled Pseudo-label Generation ~(DPG) module, designed to efficiently generate pseudo-labels by separately processing 2D and 3D attributes. This module incorporates a unique homography-based method for identifying dependable pseudo-labels in Bird's Eye View ~(BEV) space, specifically for 3D attributes. Additionally, we present a Depth Gradient Projection ~(DGP) module to mitigate optimization conflicts caused by noisy depth supervision of pseudo-labels, effectively decoupling the depth gradient and removing conflicting gradients. This dual decoupling strategy—at both the pseudo-label generation and gradient levels—significantly improves the utilization of pseudo-labels in SSM3OD. Our comprehensive experiments on the KITTI benchmark demonstrate the superiority of our method over existing approaches.
\end{abstract}    
\section{Introduction} 
Monocular 3D Object Detection (M3OD) is designed to detect objects in 3D space using a single 2D RGB image as input, playing a pivotal role in contemporary 3D perception systems across applications
\begin{figure}[h]
    \centering
    \includegraphics[width=0.9\linewidth]{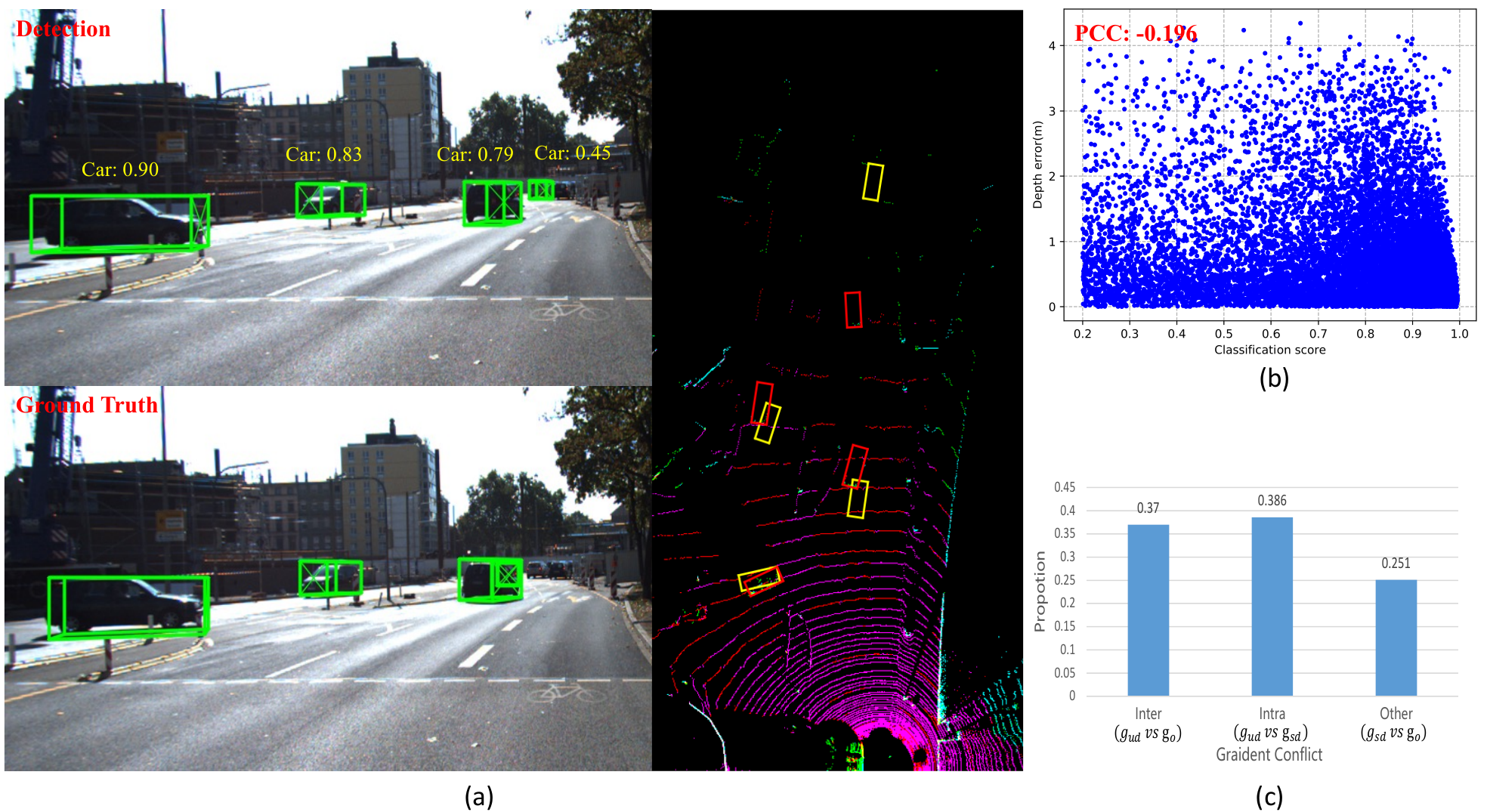}
    \caption{(a) Visualization of Pseudo-Labels and Ground Truth in Image Plane and Bird's Eye View (BEV) Plane. \textcolor{red}{Red}: Ground Truth. {\color{orange}Yellow}: Detected Bounding Boxes. (b) Statistical analysis of classification scores and depth errors~(PCC: Pearson Correlation Coefficient). (c) The proportion of different types of gradient conflicts occurring. \textbf{$g_{sd}$}: Gradient of ground truth depth loss. \textbf{$g_{ud}$}: Gradient of pseudo-label depth loss. \textbf{$g_o$}: Gradient of other attribute supervision loss. Gradient conflicts between $g_i$, $g_j$ when $\cos(g_i,g_j)<0$.}
    \label{fig:teaser}
\end{figure}
like autonomous driving and robotic navigation. The major challenge in current M3OD methods lies in their dependence on precise annotations, a labor-intensive and costly process. To overcome this obstacle, Semi-Supervised Monocular 3D Object Detection (SSM3OD) has emerged as a promising solution. It capitalizes on the abundance of readily available unlabeled images to enhance the performance of M3OD detectors. 
In line with prevalent semi-supervised learning techniques~\cite{mixmatch,fixmatch,mean-teacher,zhang2023semidetr,li2023gradient}, pseudo-labeling and consistency regularization are two kinds of widely used technology in SSM3OD~\cite{mix-teaching,mvc-monodet}. This paper specifically explores the pseudo-labeling technique within the realm of SSM3OD.

M3OD is inherently a multi-task challenge, encompassing a range of both 2D~(e.g.~classification) and 3D~(e.g.~depth) attribute predictions. We observe that there is a significant disparity between the 2D and 3D attributes. As illustrated in Fig.~\ref{fig:teaser}~(a), the detection with high confidence scores has subpar depth and orientation predictions in considerable cases. Taking the depth attribute as an example, our analysis reveals a weak correlation (PCC: -0.196) between the quality of depth prediction and the associated classification confidence, as depicted in Fig.~\ref{fig:teaser}~(b). This issue stems from the perspective projection,  which complicates the distinction of 3D attribute quality on the 2D image plane, as illustrated in Fig. ~\ref{fig:teaser}~(a). However, most existing SSM3OD methods~\cite{mvc-monodet,mix-teaching} overlook this disparity and only rely on the accuracy of 2D attributes(e.g. confidence score) to achieve pseudo-label generation, which leads to unreliable supervision for the 3D attributes.

To address this issue, we introduce a \textbf{decoupled pseudo-label generation}~(DPG) module to generate more effective pseudo-labels for both 2D and 3D attributes. Specifically, given the disparity between 2D and 3D attributes, we propose to separate the pseudo-label generation for these two types and develop a Homography-based Pseudo-label Mining~(HPM) module to generate pseudo-labels specifically for 3D attributes. Leveraging the estimated 2D-3D homography transformation, HPM transforms predictions from the 2D image plane to the 3D Bird's Eye View (BEV) plane, in which the pseudo-labels with reliable 3D attributes are iteratively identified based on the localization error. However, due to the noisy nature of the depth estimation, we observe a frequent conflict between the depth supervision derived from pseudo-labels and other reliable supervision sources~(ground truth of depth, ground truth \& pseudo-label of the attributes except depth). As illustrated in Fig.~\ref{fig:teaser}~(c), the gradient conflicts between the pseudo-label depth loss and other reliable supervision loss (represented as $g_{ud}$ vs $g_{sd}$ and $g_{ud}$ vs $g_{o}$) is more prevalent compared to conflicts within the reliable supervision ($g_{sd}$ vs $g_{o}$). Such gradient conflicts potentially undermine the utilization of reliable supervision. 

To mitigate this issue, we further develop a \textbf{depth gradient projection}~(DGP) module. This module effectively projects the conflicting depth gradient towards the principal reliable gradient, eliminating the harmful component. This adjustment ensures that the depth supervision derived from pseudo-labels is always in harmony with reliable supervision. By incorporating both the DPG and DGP modules, our Decoupling Pseudo-Labeling~(DPL) approach significantly enhances the generation and utilization of pseudo-labels for SSM3OD. We have conducted comprehensive experiments to validate the efficacy of our method on the KITTI~\cite{kitti} benchmark and achieved state-of-the-art results. Our contributions can be summarized as follows:
\par

\begin{itemize}
\item We identify and address the quality misalignment between the predictions of 2D and 3D attributes, an issue previously overlooked in existing pseudo-labeling SSM3OD methods. 
\item We introduce a decoupled pseudo-label generation module featuring a homography-based depth label mining module to generate reliable pseudo-labels for both 2D and 3D attributes.
\item We develop a depth gradient projection module to mitigate the adverse effects potentially caused by noisy depth pseudo-labels.
\item Our extensive experimental results on the KITTI benchmark demonstrate that our approach significantly surpasses all previous state-of-the-art methods.  
\end{itemize}
\section{Related Work}
\textbf{Monocular 3D Object Detection}. Monocular 3D object detection~(M3OD) aims to detect objects within a three-dimensional space utilizing solely a single camera. Existing methods in M3OD can be broadly categorized into two streams: one that relies exclusively on monocular images, and another that incorporates supplementary data sources, such as CAD models~\cite{autoshape}, dense depth map~\cite{depth_guide,ddmp_3d,patch_net,pseudo_lidar,pseudo_lidar++}, and LiDAR~\cite{monorun,monodtr,caddn, monojsg,monodtr,kinematic}. Owing to their cost-effectiveness and ease of deployment, we focus on the methods that rely solely on monocular images as input. Initial efforts in the field~\cite{m3d-rpn, smoke, fcos3d} adapted conventional 2D object detection frameworks~\cite{faster_rcnn, fcos, center-net} to incorporate 3D object detection capabilities. Studies such as MonoDLE~\cite{monodle} and PGD~\cite{pgd} have highlighted the critical challenge in M3OD: precise depth prediction. In response, numerous studies have sought to harness the synergy between 2D-3D geometric relationships~\cite{gupnet,monorcnn,rtm3d,monodde,mono_disentagle} or exploit spatial context~\cite{monopair,homoloss,extra_context,dense_contraint,monodde} to enhance depth estimation accuracy. MonoFlex~\cite{monoflex} introduces a novel depth ensemble approach, synthesizing various depth estimation techniques to elevate detection performance significantly. However, these advancements are largely contingent upon annotations with precise depth, which are labor-intensive and costly to obtain. Consequently, this research explores the potential of semi-supervised learning methodologies to alleviate the annotation burden. 

\noindent\textbf{Semi-Supervised Monocular 3D Object Detection}. Semi-supervised monocular 3D object detection~(SSM3OD) harnesses a wealth of unlabeled monocular imagery alongside a limited corpus of precisely annotated labeled monocular images to enhance monocular 3D object detection efficacy. Mix-Teaching~\cite{mix-teaching} introduces a database-oriented pseudo-labeling strategy that pastes pseudo-instances onto the background unlabeled images, thereby generating additional training samples. 
It further includes a model prediction ensemble-based pseudo-label filter to isolate high-quality pseudo-labels. However, this method does not fully address the distinct characteristics between 2D and 3D attributes, resulting in sub-optimal exploitation of 3D information in pseudo-label generation, and consequently, less effective pseudo-labels. MVC-MonoDet~\cite{mvc-monodet} focuses on the consistency regularization technique and designs a multi-view consistency strategy to exploit the depth clue in the unlabeled multi-view monocular images~(video, stereo images). Our method is orthogonal with~\cite{mvc-monodet} and focuses specifically on the relatively underexplored pseudo-labeling strategies within SSM3OD. It is noteworthy that~\cite{lidar-guided,empirical-semi} also propose pseudo-labeling methods for monocular 3D object detection. However, these methods derive pseudo-labels using LiDAR point clouds, which inherently provide accurate depth information for objects. In contrast, our method generates pseudo-labels exclusively from monocular images, without relying on supplementary LiDAR data, presenting a more challenging yet practical scenario.

\section{Preliminary}
\textbf{Problem Definition}. Given the labeled dataset as $\boldsymbol{D}_l = {(\boldsymbol{I}^l_i, \boldsymbol{y}_{i}^{l})\}_{i=1}^{N_{l}}}$, and unlabeled dataset $\boldsymbol{D}_u = \{(\boldsymbol{I}^u_i)\}_{i=1}^{N_{u}}$, where $\boldsymbol{I}^l_i$ and $\boldsymbol{I}^u_i$ denote an RGB image of labeled dataset and the unlabeled dataset respectively, and $N_{l}$ and $N_{u}$ is the corresponding data amounts. $\boldsymbol{y}_{i}^{l} = \{(c_{j}^{l}, \boldsymbol{o}_{j}^{l})\}_{j=1}^{N_{i}}$ is a list of $N_{i}$ 3D bounding box annotations for $i$-th labeled image, where $c_{j}^{l}$ is the category label and $\boldsymbol{o}_{j}^{l}$ is the 3D box label including the orientation, dimension, and location. The target of SSM3OD is to achieve monocular 3D object detection by leveraging limited labeled images with additional abundant unlabeled images. The optimization of the SSM3OD can be formulated as:
\begin{equation}
    \mathcal{L}= \mathcal{L}_{sup} + \alpha  \mathcal{L}_{unsup},
\label{eq:ssm3od_loss}
\end{equation}
where $\mathcal{L}_{sup}$ and $\mathcal{L}_{unsup}$ are the supervised loss and unsupervised loss and $\alpha$ is the loss weight and set to 1 by default.

\section{Method}
\begin{figure*}[ht]
    \centering
    \includegraphics[width=0.9\linewidth]{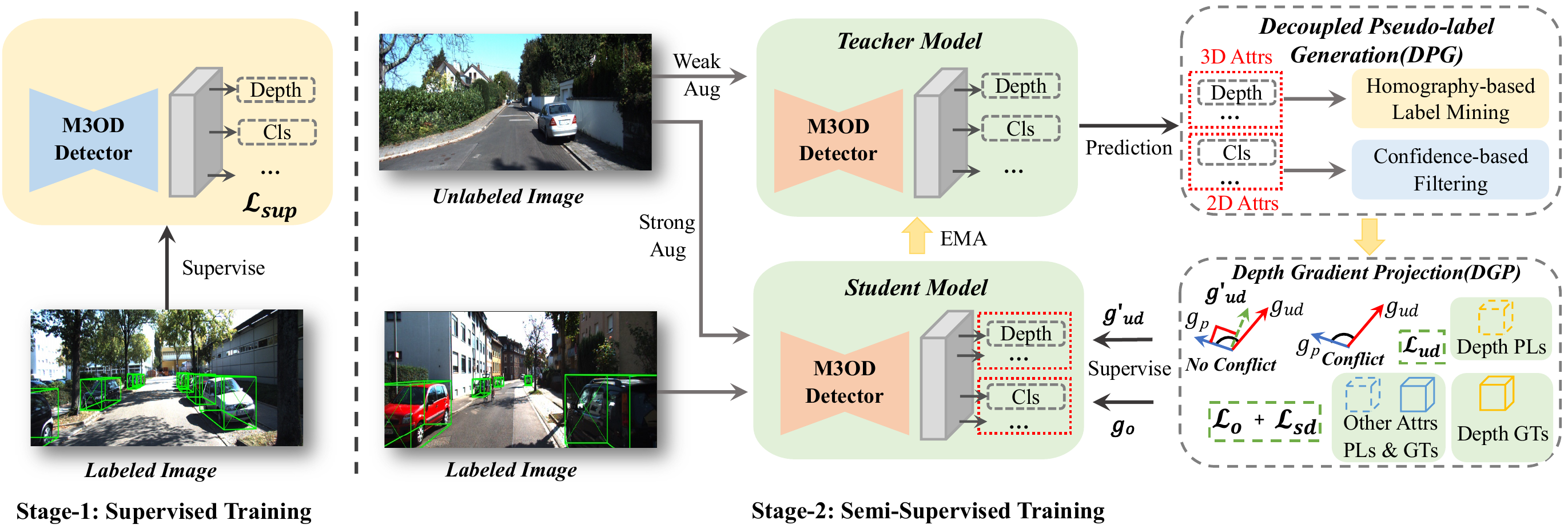}
    \caption{The overview of the \textbf{Decoupled Pseudo-Labeling}~(DPL) method for SSM3OD. We conduct semi-supervised learning based on the teacher-student framework after the supervised training stage. DPL consists of \textit{Decoupled Pseudo-label Generation}~(DPG) module and \textit{Depth Gradient Projection}~(DGP) module.
    DPG decouple the 2D and 3D attribute and generate pseudo-labels independently, with a
    \textit{Homography-based Label Mining}~(HLM) algorithm designed to generate  pseudo-labels 3D attributes by harnessing the homography transformation. DGP module utilizes a gradient projection operation to mitigate the potential negative impact of noisy depth supervision.}
    \label{fig:overview}
\end{figure*}

Fig.~\ref{fig:overview} presents an overview of our Decoupled Pseudo-Labeling (DPL) approach for Semi-Supervised Monocular 3D Object Detection (SSM3OD). It employs the classic teacher-student framework~\cite{mean-teacher}, where a teacher and student network are involved, both of which are initialized by a supervised pre-trained model. The teacher model generates pseudo boxes on unlabeled images, while the student model is trained with labeled images with ground truth annotation and unlabeled images with pseudo labels. The teacher model is iteratively updated from the student model using the exponential moving average (EMA) strategy. Our DPL method integrates two key modules: Decoupled Pseudo-label Generation (DPG) and Depth Gradient Projection (DGP), to enhance the effectiveness of pseudo-label utilization in SSM3OD.

\subsection{Decoupled Pseudo-label Generation}~\label{decoupled_pseudo_label_generation} 
Given the fact that the prediction quality of 2D and 3D attributes in SSM3OD is not aligned, we propose to decouple the pseudo-label generation process for these two types of attributes. We classify object category, 2D size, and the projected 3D center as 2D attributes, and depth, 3D size, and orientation as 3D attributes. For 2D attributes, the classification confidence threshold $\theta_c$ is used to filter pseudo-labels, following~\cite{mvc-monodet}. For the 3D attributes, we introduce a novel approach that leverages the homography geometric relationship for 3D attribute pseudo-label generation. This forms the basis of our Decoupled Pseudo-label Generation (DPG), the unsupervised loss can be formulated as:
\begin{equation}
    \mathcal{L}_{unsup}= \mathcal{L}_{2D}(x_{2D}, y_{2D}) + \alpha  \mathcal{L}_{3D}(x_{3D}, y_{3D}),
\label{eq:ssm3od_unsup_loss}
\end{equation}
where $x_{2D}$, $x_{3D}$ are 2D and 3D attributes prediction of model output, and $y_{2D}$, $y_{3D}$ are the two groups of pseudo-labels for 2D and 3D attributes, respectively. The loss functions $\mathcal{L}_{2D}$ and $\mathcal{L}_{3D}$ are consistent with MonoFlex~\cite{monoflex}. With the decoupled design, both 2D and 3D attributes can be supervised with more effective pseudo-labels.\par
\noindent\textbf{Homography-based Pseudo-Label Mining}\label{homography-depth-pseudo-label-mining}
Due to the perspective projection, accurately gauging the quality of 3D attribute predictions for bounding boxes on the image plane poses a challenge. In response, we have crafted a unique homography-based depth pseudo-label mining module. This module's pivotal feature is the transposition of predictions from the 2D image plane to the Bird's Eye View (BEV) plane using homography transformation. This shift significantly improves the precision of assessing 3D attribute predictions, such as depth and orientation.

\begin{figure}[ht]
    \centering
    \includegraphics[width=0.9\linewidth]{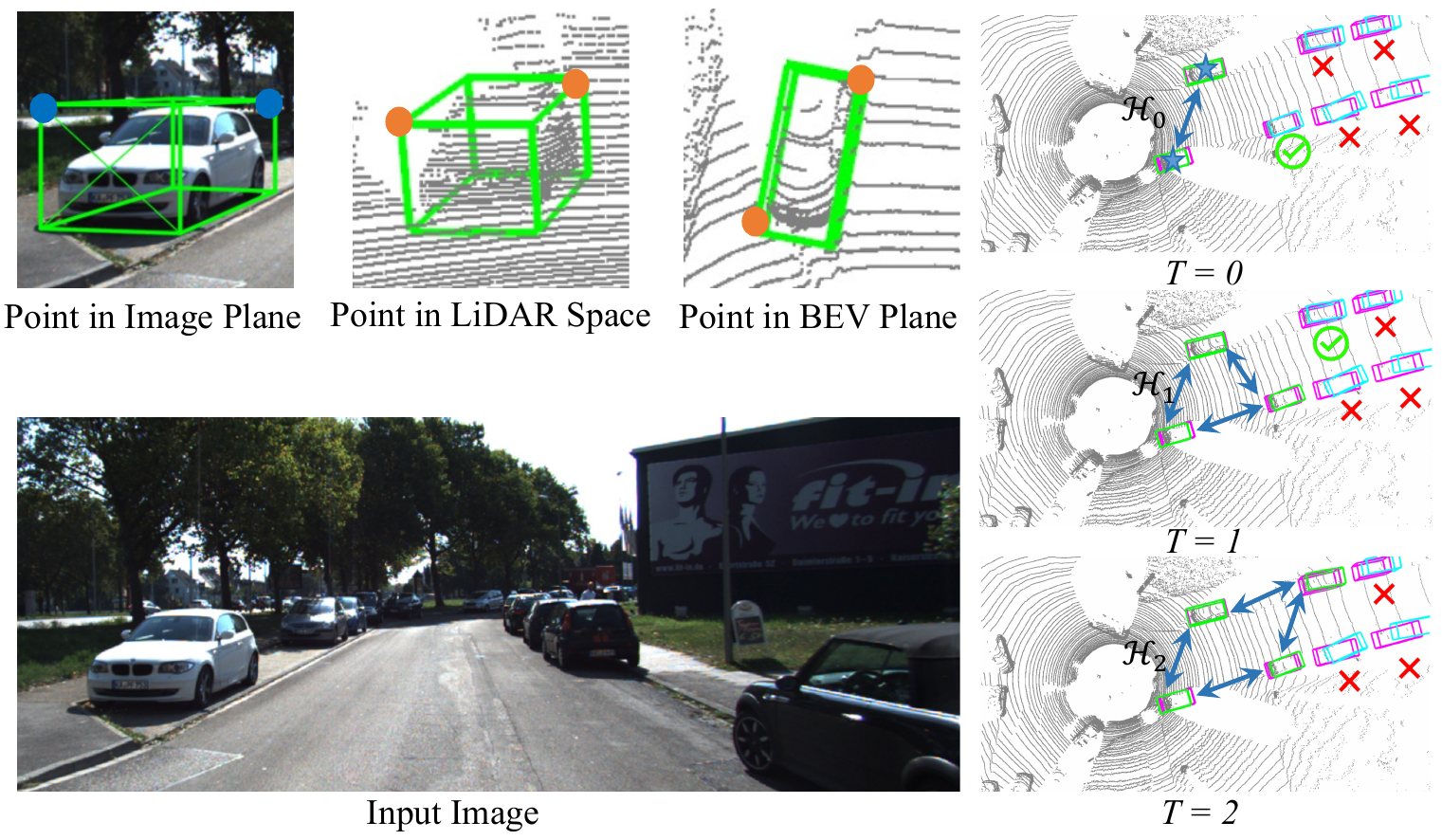}
    \caption{Illustration of HPM module. The homography transformation describes the coordinate mapping between the image plane and the BEV plane. Starting with the initial pseudo-labels, we iteratively estimate the homography matrix $H_{i}$ and search the reliable pseudo-labels in BEV space. {\color{green}Green} Box: The selected Pseudo-Labels. {\color{cyan}Cyan} Box: Model Prediction. {\color{purple}Purple} Box: Ground Truth.}
    \label{fig:homography}
\end{figure}

\noindent\textit{Homography Transformation}: Generally, let the coordinates of a point on the \textbf{ground surface} as $I =(u, v)$ in the 2D Image plane and $B = (x^b, y^b)$ in the BEV plane as shown in Fig.~\ref{fig:homography}. The transformation between homogeneous coordinates $(u, v, 1)$ and  $(x^b, y^b, 1)$ can be describe by a homography matrix $\boldsymbol{M} \in \mathbb{R}^{3 \times 3}$:
\begin{equation}
[x^b, y^b, 1]^T = \boldsymbol{M}[u, v, 1]^T.
\label{eq:homo}
\end{equation}
The homography transformation ~\cite{homography,3d_homography} is a geometric relationship between two coordinate systems of 2D and 3D plane. Therefore, with the flat ground assumption~\cite{homoloss}, different objects within an image will share the same homography matrix. \par 
\noindent\textit{Iteratively Pseudo-Label Mining}: We develop an iterative pseudo-label mining algorithm to acquire reliable 3D attribute pseudo-labels, as detailed in \ref{homography-depth-pseudo-label-mining}. This algorithm uses the deviation from a consistent homography transformation as a measure of 3D attribute prediction quality.   

Specifically, it initially selects pseudo-labels with relatively accurate 3D attribute predictions to estimate the initial homography matrix reliably. To assist in this initial generation of pseudo-labels, we employ the depth prediction uncertainty $\sigma$ from the Laplacian aleatoric uncertainty loss~\cite{monopair}, as defined in Eq.~\ref{eq:depth_uncertainty_loss}:
\begin{equation}
    \mathcal{L}_{d e p t h}=\frac{\sqrt{2}}{\sigma}\left\|d_{g t}-d_{\text {pred }}\right\|_{1}+\log (\sigma),
\label{eq:depth_uncertainty_loss}
\end{equation}
where $d_{gt}$ is the ground truth depth and $d_{\text {pred }}$ is the predicted depth of an object, $\sigma$ is known as the predicted depth uncertainty to weight the prediction. We select the pseudo-labels with $\sigma < \theta_u$ as the initial pseudo-labels.\par
 
Then, we select the bottom corner points and bottom center points ~(5 points) of each initial 3D pseudo-bounding box as candidate points to estimate the homography matrix. The coordinates of these points in the image plane $\tilde{I} = (\tilde{u}, \tilde{v})$ are directly predicted by the teacher model~\cite{monoflex,monodle}. We then estimate the coordinates of these points in the BEV plane. Specifically, we first get the positions of these points in the camera coordinate system $(\tilde{x}, \tilde{y}, \tilde{z})$ by the local transformations \cite{monojsg} which is estimated from the dimensions, orientations, and positions of the centers of the 3D boxes.

Then the coordinate in the lidar coordinate system can be obtained by projecting with the inverse of extrinsic matrices $[\mathbf{R} | \mathbf{T}]$,
\begin{equation}
[\tilde{x}^b, \tilde{y}^b, \tilde{z}^b]^T = [\mathbf{R} | \mathbf{T}]^{'}[\tilde{x}, \tilde{y}, \tilde{z}]^T.
\label{eq:extrinsic}
\end{equation}
Therefore, the coordinates of these points in the BEV plane are $\tilde{B} = (\tilde{x}^b, \tilde{y}^b)$.
Denote the candidate points coordinates in the image plane and the BEV plane of $N$ objects, as $\tilde{C}_I \in \mathbb R^{2 \times 5N} $  and  $\tilde{C}_B \in \mathbb R^{2 \times 5N}$, the homography transformation $\tilde{\boldsymbol{M}}$ can be derived by solving Eq.\ref{eq:homo} with Direct Linear Transform~(DLT)~\cite{dlt}. 

Finally, we apply the estimated homography matrix $\tilde{\boldsymbol{M}}$ to the candidate points of the predicted bounding boxes that have not been chosen as pseudo-labels yet to get their desired coordinates in BEV space:
\begin{equation}
    \hat{B} = [\hat{x}^b, \hat{y}^b, 1]^T = \tilde{\boldsymbol{M}} [\tilde{u}, \tilde{v}, 1]^T .
\end{equation}
Ideally, the BEV coordinated obtained by the homography transformation $\hat{c}^b = (\hat{x}^b, \hat{y}^b)$ should be the same as $\tilde{c}_t^b =
 (\tilde{x}^b, \tilde{y}^b)$ estimated from the model prediction via Eq.\ref{eq:extrinsic}. However, the poorly predicted 3D attributes ~(e.g. depth, orientation) would result in the deviation. Therefore, this deviation can serve as the proxy for the prediction quality of these attributes, and we select the prediction satisfying $||\hat{c}^b - \tilde{c}_t^b ||_2 < \theta_h$ as the eligible pseudo-labels, where $\theta_h$ is the pre-defined threshold. The newly obtained pseudo-labels are then appended to the previously acquired pseudo-labels to start a new iteration of homography matrix estimation and pseudo-label filtering as shown in Fig.~\ref{fig:homography}. The iterative process continues until either no new pseudo-labels are obtained or the predefined maximum iteration limit, denoted as $t_{max}$, is reached. Please refer to the Appendix for the complete algorithm.
 
\subsection{Depth Gradient Projection}
Due to the inherent limitations of depth estimation from monocular images, the depth supervision from pseudo-labels will be inevitably noisy and can cause optimization conflict~(i.e. conflicted gradient direction) with reliable supervision. Concretely, we split the total loss into pseudo-label depth loss $\mathcal{L}_{ud}$, ground truth depth loss $\mathcal{L}_{sd}$ and other attributes loss $\mathcal{L}_{o}$~(including both labeled and unlabeled loss). Their gradient are denoted as $g_{ud}(\theta)$, $g_{sd}(\theta)$ $g_{o}(\theta)$, respectively. Generally, since the pseudo-labels of the attributes except for depth can be estimated with reasonable accuracy than depth~\cite{pgd}, $\mathcal{L}_{sd}$ and $\mathcal{L}_{o}$ can be regarded more reliable than $\mathcal{L}_{ud}$. We check the optimization conflicts between different supervision as presented in Fig.~\ref{fig:teaser}. It clearly shows that the gradient from depth pseudo-labels conflicts with reliable supervision more frequently.

To address this issue, we develop a simple depth gradient projection module to eliminate the possible negative impact of noisy depth supervision from the gradient perspective. 

Concretely, given that $g_{sd}$ and $g_{o}$ conflict less frequently, we combine them together and treat them as the optimization principle gradient $g_{p} = g_{o} + g_{sd}$, which stands for optimization direction of reliable supervision. Then, we project the $g_{ud}(\theta)$ to the normal vector of gradient $g_{p}(\theta)$ when the conflict occurs:
\begin{equation}
\footnotesize
\label{eq}
 g_{ud}^{'}(\theta)=\left\{
\begin{aligned}
g_{ud}(\theta) - \frac{g_{ud}(\theta) g_{p}(\theta) }{||g_{p}(\theta)||_{2}^{2} }. g_{p}(\theta)&, & if\ cos(g_{ud},g_p)<0, \\
g_{ud} & , & otherwise
\end{aligned}
\right.
\end{equation}

The obtained gradient $g_{ud}^{'}(\theta)$ thus has no conflicted gradient component with $g_{p}(\theta)$. Equipped with this module, the noisy unsupervised depth loss is always guaranteed to share common interests with reliable supervision and deliver an equilibrium optimization target.
\section{Experiments}
\subsection{Experimental Setup}
\textbf{Dataset and Metrics}. \textbf{KITTI} dataset~\cite{kitti} is the standard dataset for M3OD, providing 7,481 images for training and 7,518 images for testing. Following the common practice~\cite{kitti_split}, the training set is further split into 3,712 training samples and 3,769 validation samples. For the unlabeled data, we select the \textit{completely unlabeled video sequence in the KITTI raw data that does not overlap with the video sequence of the training and validation split}.  This results in approximately 35K unlabeled images, which are utilized as our unlabeled dataset. We report the evaluation results on the validation and test set based on $AP|_{R_{40}}$.

\noindent\textbf{Implementation Details}. We choose  MonoFlex \cite{monoflex}, a representative M3OD detector, to evaluate the effectiveness of our method. All experiments are conducted using the official code provided by the author. We first pre-train the model using labeled data and then perform end-to-end semi-supervised learning using both labeled and unlabeled data. Each unlabeled image has weak and strong augmented versions, which are sent to the teacher and student network respectively. The strong augmentation includes random horizontal flips, photometric distortion, random gray, and random Gaussian blur, while the weak augmentation only involves random horizontal flips. During pseudo-label generation with the outputs of the teacher network, we first filter out predictions that are background with a classification score threshold of 0.2. The $\theta_c$, $\theta_u$, and $\theta_h$ in the DPG module are set to 0.4, 0.1, and 2.0 respectively, and $t^{max}$ is set to 10. More implementation details are in the Appendix.

\begin{table*}[t]

\centering
\caption{Comparision with state-of-the-art (SOTA) Methods. We present the evaluation results of \textbf{‘Car’ category in the KITTI test set}. \textbf{†} denotes our reproduction results. For fair comparisons, we train Mix-Teaching with the same data volume as our method.}
\label{main_results_test}
\begin{tabular}{c|c|c|c|c|c|c|c}
\hline 
\multirow{2}{*}{ Method }  & \multirow{2}{*}{ Extra Data } & \multicolumn{3}{c}{ Test $A P_{3 D}|R_{40}$} & \multicolumn{3}{|c}{ Test $A P_{B E V}|R_{40}$} \\

\cline { 3 - 8 }& & Easy & Mod. & Hard & Easy & Mod. & Hard \\
\hline 
PatchNet &  \multirow{3}{*}{ Depth } & 15.68 & 11.12 & 10.17 & 22.97 & 16.86 & 14.97 \\
 D4LCN &  & 16.65 & 11.72 & 9.51 & 22.51 & 16.02 & 12.55  \\
 DDMP-3D &  & 19.71 & 12.78 & 9.80 & 28.08 & 17.89 & 13.44 \\
\hline Kinematic3D & Multi-frames & 19.07 & 12.72 & 9.17 & 26.69 & 17.52 & 13.10  \\
\hline MonoRUn &  \multirow{3}{*}{ LiDAR } & 19.65 & 12.30 & 10.58 & 27.94 & 17.34 & 15.24  \\
 CaDDN &  & 19.17 & 13.41 & 11.46 & 27.94 & 18.91 & 17.19 \\
 MonoDTR & & 21.99 & 15.39 & 12.73 & 28.59 & 20.38 & 17.14 \\
\hline AutoShape & CAD & 22.47 & 14.17 & 11.36 & 30.66 & 20.08 & 15.59  \\
\hline SMOKE & \multirow{8}{*}{ None } & 14.03 & 9.76 & 7.84 & 20.83 & 14.49 & 12.75  \\
 MonoPair &  & 13.04 & 9.99 & 8.65 & 19.28 & 14.83 & 12.89 \\
RTM3D &  & 13.61 & 10.09 & 8.18 & - & - & -  \\
PGD &  & 19.05 & 11.76 & 9.39 & 26.89 & 16.51 & 13.49 \\
 MonoRCNN &  & 18.36 & 12.65 & 10.03 & 25.48 & 18.11 & 14.10 \\
Zhang et al. DLE &  & 20.25 & 14.14 & 12.42 & 28.85 & 17.72 & 17.81 \\
GUPNet &  & 20.11 & 14.20 & 11.77 & - & - & - \\
HomoLoss &  & 21.75 & 14.94 & 13.07 & 29.60 & 20.68 & 17.81 \\
\hline Mix-Teaching & Unlabeled & 21.88 & 14.34 & 11.86 & 30.52 & 19.51 & 16.45  \\
MVC-MonoDet & Unlabeled & 22.13 & 14.56 & 12.09 & 31.62 & 20.11 & 17.21  \\
\hline

 MonoFlex† &  None & 19.23 & 12.57 & 10.73 & 26.83 & 17.94 & 15.16  \\
\rowcolor{gray!40}DPL$_{FLEX }$ & Unlabeled & \textbf{24.19} &  \textbf{16.67} & \textbf{13.83} & \textbf{33.16} & \textbf{22.12} & \textbf{18.74} \\
\textbf{\emph{Improvement}} &  v.s. Baseline & \textbf{\textcolor{red!70!black}{+4.96}} & \textbf{\textcolor{red!70!black}{+4.10}} & \textbf{\textcolor{red!70!black}{+3.10}} & \textbf{\textcolor{red!70!black}{+6.33}}& \textbf{\textcolor{red!70!black}{+4.18}} & \textbf{\textcolor{red!70!black}{+3.58}}  \\

\hline
\end{tabular}
\end{table*}

\begin{table*}[h]
\centering
\caption{Performance of \textbf{`Car' category in the KITTI validation set} based on MonoFlex under different amounts of unlabeled data. We randomly chose 5K, 15K, and 25K data from the whole 35K KITTI unlabeled set as the unlabeled data.}

\label{validation_num}
\setlength\tabcolsep{3pt} 
\begin{tabular}{c|ccc|ccc|ccc|ccc}
\hline \multirow{3}{*}{ Methods } & \multicolumn{12}{|c}{ Val, $A P_{3 D}|R_{40}$} \\
\cline{2-13}
 &\multicolumn{3}{|c}{ 5K unlabel}& \multicolumn{3}{|c}{ 15K unlabel} &\multicolumn{3}{|c}{ 25K unlabel} &\multicolumn{3}{|c}{ 35k unlabel} \\
\cline { 2 - 13}  & Easy & Mod & Hard& Easy & Mod & Hard & Easy & Mod & Hard  & Easy & Mod & Hard\\
\hline
MonoFlex † &22.80 &17.51& 14.90&22.80 &17.51& 14.90&22.80 &17.51& 14.90&22.80 &17.51& 14.90 \\
\rowcolor{gray!40}DPL$_{FLEX }$& \textbf{24.61}&  \textbf{18.76} &  \textbf{16.39} &  \textbf{25.23} &  \textbf{18.86} &  \textbf{16.47} &  \textbf{26.05} &  \textbf{19.22}  & \textbf{16.84} &  \textbf{26.51} & \textbf{19.84}  &  \textbf{17.13}  \\
\textbf{\emph{Improvement}} &\textbf{\color{blue}{+1.81} }& \textbf{\color{blue}{+1.25}}&\textbf{\color{blue}{ +1.49}} &\textbf{\color{blue}{+2.43} }& \textbf{\color{blue}{+1.35}}&\textbf{\color{blue}{ +1.57}} & \textbf{\color{blue}{+3.25}}&\textbf{\color{blue}{ +1.71}}  &\textbf{\color{blue}{+1.94}} & \textbf{\color{blue}{+3.71}} &\textbf{\color{blue}{+2.33}}  & \textbf{\color{blue}{+2.23}}  \\
\hline
\end{tabular}
\end{table*}

\begin{table*}[h]
\centering
\caption{Performance of \textbf{‘Car’ category in the KITTI validation set} based on MonoFle under different labeled ratios. We randomly chose 10\%, 50\%, and 100\% of KITTI train split as the labeled data.}
\label{different_label_ration}
\begin{tabular}{c|ccc|ccc|ccc}
\hline \multirow{3}{*}{ Methods } & \multicolumn{9}{|c}{ Val, $A P_{3 D}|R_{40}$} \\
\cline{2-10}
 & \multicolumn{3}{|c}{ 10\%} &\multicolumn{3}{|c}{ 50\%} &\multicolumn{3}{|c}{ 100\%} \\
\cline { 2 - 10} & Easy & Mod & Hard & Easy & Mod & Hard  & Easy & Mod & Hard\\
\hline
MonoFlex † & 4.77& 3.90 & 3.29 & 18.85 & 14.81  &12.48 &22.80& 17.51 &14.90 \\
\rowcolor{gray!40}DPL$_{ {FLEX }}$ &\textbf{10.25} & \textbf{8.19} & \textbf{7.09} & \textbf{21.33} & \textbf{16.42}  &\textbf{14.57 }&\textbf{ 26.51} &\textbf{19.84}  &\textbf{ 17.13}  \\
\textbf{\emph{Improvement}} &\textbf{\color{blue}{+5.48}} & \textbf{\color{blue}{+4.29}}&\textbf{\color{blue}{+3.8}} & \textbf{\color{blue}{+2.48}}&\textbf{\color{blue}{+1.61}}  &\textbf{\color{blue}{+2.09}} & \textbf{\color{blue}{+3.71} }&\textbf{\color{blue}{+2.33}}  & \textbf{\color{blue}{+2.23}}  \\
\hline
\end{tabular}
\end{table*}

\begin{table*}[h]
\centering
\caption{Performance of \textbf{`Car' category in the KITTI validation set} under different base detectors. * Note that the provided code of MonoDETR is only an intermediate version (not complete) which is confirmed officially by the authors. We also do not reproduce the results reported in their paper.}

\setlength\tabcolsep{3pt} 
\begin{tabular}{c|ccc|ccc|ccc|ccc}
\hline 
\label{different_mono_detectors}
\multirow{3}{*}{ Methods } & \multicolumn{12}{|c}{ Val, $A P_{3 D}|R_{40}$} \\
\cline{2-13}
 &\multicolumn{3}{|c}{ MonoDLE}& \multicolumn{3}{|c}{ PGD} &\multicolumn{3}{|c}{ GUPNet} &\multicolumn{3}{|c}{ MonoDETR*} \\
\cline { 2 - 13}  & Easy & Mod & Hard& Easy & Mod & Hard & Easy & Mod & Hard  & Easy & Mod & Hard\\
\hline
Sup Baseline &17.25& 13.87 & 11.83& 19.27& 13.22 & 10.64 & 22.76 & 16.46  &13.72 &26.95& 18.87 &15.52 \\
\rowcolor{gray!40}DPL & \textbf{19.31}&  \textbf{15.67} &  \textbf{13.72} &  \textbf{21.34} &  \textbf{15.34} &  \textbf{12.49} &  \textbf{24.48} &  \textbf{18.51}  & \textbf{14.89} &  \textbf{28.12} & \textbf{20.81}  &  \textbf{17.37}  \\
\textbf{\emph{Improvement}} &\textbf{\textcolor{red!70!black}{+2.06} }& \textbf{\textcolor{red!70!black}{+1.80}}&\textbf{\textcolor{red!70!black}{ +1.50}} &\textbf{\textcolor{red!70!black}{+2.07} }& \textbf{\textcolor{red!70!black}{+2.12}}&\textbf{\textcolor{red!70!black}{ +1.85}} & \textbf{\textcolor{red!70!black}{+1.72}}&\textbf{\textcolor{red!70!black}{ +2.05}}  &\textbf{\textcolor{red!70!black}{+1.17}} & \textbf{\textcolor{red!70!black}{+1.17}} &\textbf{\textcolor{red!70!black}{+1.94}}  & \textbf{\textcolor{red!70!black}{+1.85}}  \\
\hline
\end{tabular}
\end{table*}
\subsection{Main Results}
 Tab.~\ref{main_results_test} presents the results on the KITTI test set. It shows that by incorporating the proposed DPL for SSM3OD, our method significantly boosts the performance of the base detector. In particular, our method boosts the performance of MonoFlex with \textbf{+4.10} and \textbf{+4.18} on $AP_{3D}$ and $AP_{BEV}$, respectively. Moreover, based on MonoFlex, our method surpasses all existing SSM3OD methods by a large margin and achieves a new state-of-the-art performance across all fully supervised and semi-supervised methods. Specifically, integrating our method into the MonoFlex, our method outperforms Mix-Teaching by \textbf{+2.33} $AP_{3D}$  and \textbf{+2.61} $AP_{BEV}$, and exceeds the performance of MVC-MonoDet by \textbf{+2.11} and \textbf{+2.01} on $AP_{3D}$ and $AP_{BEV}$.

\subsection{Ablation Study}
\textbf{Components Effectiveness}. We ablate the effects of decoupled pseudo-label generation~(DPG) and depth gradient projection~(DGP) module. The results are presented in Tab.\ref{ablation_components_effectivenss}. We start from the \textbf{SSM3OD Baseline} that utilizes the classification confidence threshold 0.6 to filter the pseudo-labels for both 2D and 3D attributes. It clearly shows that integrating the decoupled pseudo-label generation(DPG) module improves the performance by \textbf{1.21} on Car $AP_{3D}$(Mod.) without bells and whistles. It demonstrates the importance of decoupling the pseudo-labels generation process for 2D and 3D attributes. Further applying the depth gradient projection~(DGP) module to eliminate potential gradient conflict leads to a \textbf{0.62} Car $AP_{3D}$(Mod.) improvement and reaches 19.85 $AP_{3D}$~(Mod.), which is \textbf{1.83}  $AP_{3D}$ higher than the baseline.

\begin{table}[h]
\setlength\tabcolsep{2pt} 
\caption{Ablation of the effectiveness of proposed components. The experiments were conducted on the KITTI validation set with MonoFlex. Car category's performance $AP_{3D}/AP_{BEV}| R_{40}$ of $IoU=0.7$ is reported}

\begin{tabular}{cc|ccc}
\hline  DPG & DGP & Easy & Mod. & Hard \\
\hline  -  & - & $23.13 / 30.14$ & $18.02 / 23.83$ & $15.24 / 20.18$ \\
  $\surd$ & - & $26.24 / 34.91$ & $19.23 / 25.64 $ & $17.04 / 22.60$ \\
 $\surd$ &$\surd$ & $\bm{26.51 / 35.02}$ & $\bm{19.85 / 26.37}$ & $\bm{17.13} / \bm{23.08}$ \\
\hline
\end{tabular}
\vspace{-0.6cm}
\label{ablation_components_effectivenss}
\end{table}
\par

\begin{figure*}[ht]
    \centering
    \setlength{\abovecaptionskip}{0.cm}
    \includegraphics[width=0.99\linewidth]{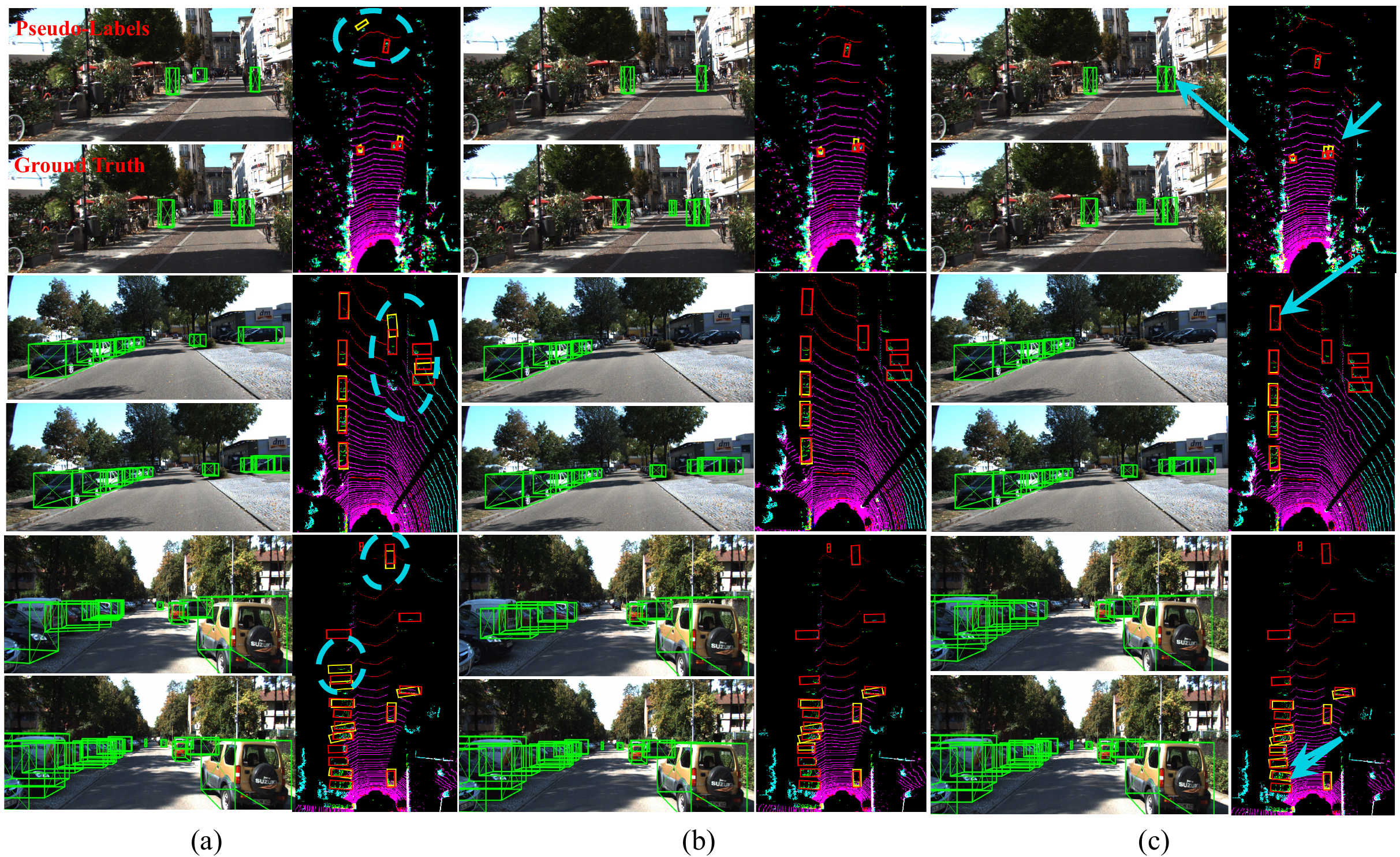}
    \caption{Visual comparison of pseudo-labels generated among different pseudo-label generation strategies. (a) Pseudo labels generated with classification confidence threshold 0.6. (b) Pseudo labels generated by initial depth prediction uncertainty filtering in HPM. (c) Pseudo labels after HPM algorithm. Red Box: Ground truth. Yellow Box: Pseudo-Labels. Cyan dashed circles: The confident yet depth-deviated pseudo-labels. Cyan arrows: The pseudo-labels discovered through homography-based mining.}
    \label{fig:visualization}
\end{figure*}

\noindent\textbf{Performance on More Base Detectors}. Beyond the MonoFlex, we conducted experiments on the KITTI validation dataset with other monocular 3D object detectors, including MonoDLE\cite{monodle}, PGD\cite{pgd}, GUPNet\cite{gupnet}, and MonoDETR\cite{monodetr}. We specifically omitted FCOS3D from this study, as its performance on the KITTI dataset is suboptimal as acknowledged by the authors. The results of these experiments are detailed in Tab.\ref{different_mono_detectors}, showing consistent and substantial performance improvements across the various base detectors. These findings underscore the strong adaptability to different monocular detectors of our method.

\noindent\textbf{Effect of Labeled and Unlabeled Data Volume}. We present the impact of labeled and unlabeled data volume on the performance of DPL in Tab.\ref{different_label_ration} and Tab.\ref{validation_num}. Our approach consistently enhances the performance of MonoFlex across various volumes of labeled data. Particularly, our method showcases significant benefits in scenarios where labeled data is scarce. For instance, we observe a substantial performance boost of \textbf{+4.29} in $AP_{3D}|40$ when only 10\% of the labeled training images are available. These results highlight the superiority of our method in effectively leveraging limited labeled data. Furthermore, as the volume of unlabeled data increases, DPL showcases a more pronounced improvement in performance. This underscores the scalability of our method, highlighting its ability to leverage larger amounts of unlabeled data effectively.

\begin{figure}[ht]
    \centering
     \setlength{\abovecaptionskip}{0.cm}
    \includegraphics[width=0.9\linewidth]{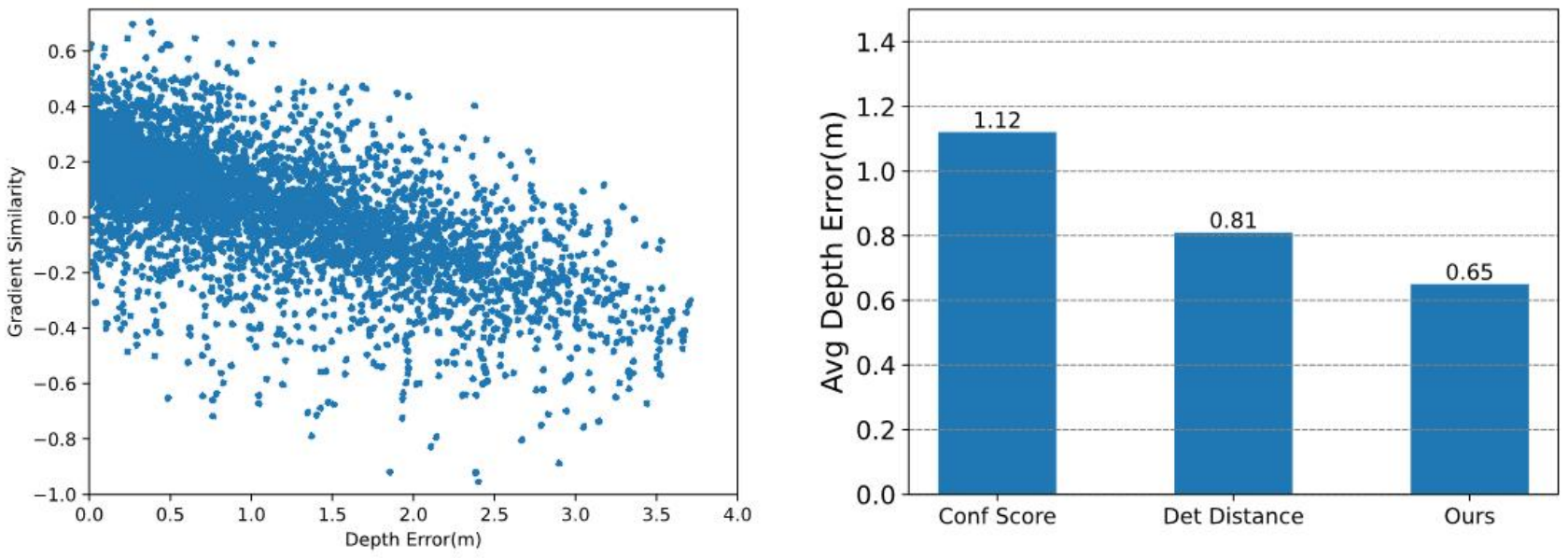}
    \caption{Left: The correlation between the gradient similarity $cos(g_{ud},g_{p})$ and the depth error. Right: The average depth error of the pseudo-labels obtained in different ways.}
    \label{fig:statistics}
\end{figure}

\noindent\textbf{Analysis of Decoupled Pseudo-label Generation}. We ablate different ways for pseudo-label generation in Tab.\ref{ablate_dpg}. It clearly shows that utilization of the classification confidence threshold(thr=0.6) for pseudo-label generation only brings limited improvement~(+0.51 $AP_{3D}$). This is attributed to its poor ability to reflect the prediction quality of 3D attributes, especially depth, leading to noisy depth pseudo-labels with large depth prediction errors as presented in Fig.\ref{fig:statistics}. As reported by \cite{monodle}, the model exhibits reasonable performance in predicting objects at close range but has limitations in predicting objects at a distance. Therefore, we generate the pseudo-labels by only retaining the prediction with a detection distance of less than 45m as suggested by \cite{monodle}. As presented in the third row of Tab.\ref{ablate_dpg}, a notable performance boost~(+1.01 $AP_{3D}$ for moderate) is observed. Nevertheless, completely disregarding pseudo-labels beyond a certain distance can impede the model's ability to detect objects that are located far away from the ego-car. This limitation is supported by the marginal improvement of 0.36 in the detection of hard category objects, compared to the use of confidence thresholding.  In contrast, our DPG leverages the geometric relationship between the 2D and 3D space through homography transformation, enabling us to derive more effective pseudo-labels with more accurate depth from the more distinguishable BEV plane as proved in Fig.\ref{fig:statistics}. By incorporating these pseudo-labels for the supervision of both 2D and 3D attributes, we ultimately achieve significantly improved performance. By further decoupling the supervision of 2D and 3D attributes and generating pseudo-labels for 2D attributes via confidence thresholding, we are able to harness the potential of pseudo-labels with accurate 2D attribute prediction but poor 3D attribute prediction. This further enhances the performance of our approach. These results clearly highlight the significance of separately processing the 2D and 3D attributes during the pseudo-labeling process.

\noindent\textbf{Visualization of Decoupled Pseudo-labeling}. We visualize the pseudo-labels generated by the classification confidence thresholding and DPG in Fig.\ref{fig:visualization}. Our DPG first selects the pseudo-labels via depth prediction uncertainty, which leads to the initial pseudo-labels with accurate depth. Subsequently, the homography-based pseudo-label mining further identifies additional pseudo-labels with reasonable depth and orientation predictions. In contrast, the pseudo-labels generated solely by confidence thresholding tend to be noisy, as they often include pseudo-labels with high confidence but inaccurate depth estimations.

\begin{table}
\footnotesize
\caption{Effectiveness of different strategies to generate the pseudo-labels for SSM3OD with MonoFlex. \textbf{cls confidence}: Filter the pseudo-labels with a classification confidence threshold 0.6. \textbf{det distance}: Generate the pseudo-labels by retaining the prediction with a detection distance of no more than 45m. \textbf{DPG w.o decouple}: Take the pseudo-labels generated by homography label mining for both 2D and 3D attributes supervision.} 
    \label{ablate_dpg}
    \centering
    \begin{tabular}{c|c|c|c}
    \hline
    \multirow{2}{*}{Strategy} & \multicolumn{3}{|c}{Val, $AP_{3D}|R_{40}$} \\
    \cline{2-4}
    & Easy & Mod & Hard \\
    \hline
    sup baseline & 22.80 & 17.51 & 14.90 \\
    \hline
    cls confidence & 23.13 & 18.02 & 15.24 \\
    det distance & 23.47 & 18.52 & 15.60 \\
    DPG w.o decouple & $25.66$ & $19.04$ & $16.24$ \\
      \hline
  
 DPG& $\mathbf{26.24}$ & $\mathbf{19.23}$ & $\mathbf{17.04}$ \\
    \hline
    \end{tabular}
\end{table}
\noindent\textbf{Analysis of Depth Gradient Projection} We conducted an analysis to examine the connection between the gradient similarity of $g_{ud}$ and $g_{p}$, and the quality of depth prediction. Our findings indicate a clear correlation between the depth error and the gradient similarity between the unsupervised depth gradient and the principal gradient. As presented in the left of Fig.\ref{fig:statistics}, it is evident that samples with larger deviations result in a more pronounced gradient conflict with reliable supervision. This further emphasizes the significance of our depth gradient projection module in mitigating the adverse effects of noisy pseudo-labels.

\section{Conclusion} 
In this work, we introduced a decoupled pseudo-labeling approach for Semi-Supervised Monocular 3D Object Detection (SSM3OD), designed to optimize the use of pseudo-labels more effectively. This approach features a decoupled pseudo-label generation module, incorporating a homography-based pseudo-label mining algorithm to efficiently provide reliable pseudo-labels for both 2D and 3D attributes. Additionally, we developed a depth gradient projection module to mitigate the adverse effects of noisy depth supervision. Comprehensive evaluations on the KITTI benchmark validate the effectiveness of our proposed method, demonstrating its superior performance in SSM3OD.

\section*{Acknowledgments}
This work was supported in part by the National Natural Science Foundation of China (NO.~62322608), in part by the CAAI-MindSpore Open Fund, developed on OpenI Community, in part by the Open Project Program of the Key Laboratory of Artificial Intelligence for Perception and Understanding, Liaoning Province (AIPU, No.~20230003). 

{
    \small
    \bibliographystyle{ieeenat_fullname}
    \bibliography{main}
}

\clearpage
\appendix

\section{Geometric Relationship in Monocular 3D Object Detection}
In contrast to 2D object detection, detecting objects in 3D space often involves a wide range of geometric relationships, and these relationships can be utilized to enhance visual perceptual performance significantly\cite{yang2023geometry}. The primary geometric prior between 2D and 3D space is the pinhole model that describes the correspondence between the location of 3D points and 2D points. With the pinhole camera model, the mapping of a 3D point $P$ with $\bm{c^\omega}=(x,y,z)$ location under the LiDAR coordinate system and its corresponding 2D location $\bm{c^i}=(u,v)$ within image can be described as:

\begin{equation}
z[u \enspace v \enspace 1]^{T}=\mathbf{K}\cdot[\mathbf{R} | \mathbf{T}]\cdot [x \enspace y \enspace z]^{T}
\end{equation}

where matrix $K$ is the \textbf{camera intrinsic matrix}. $z$ is the depth value at point $P$, $R$ and $T$ is the rotation and transformation matrix of the \textbf{camera extrinsic matrix}. In this equation,  the camera extrinsic matrix $[\mathbf{R} | \mathbf{T}]$ is responsible for the transformation between the LiDAR coordinate system and the camera coordinate system. The camera intrinsic matrix $K$ is used to transform the point from the camera coordinate to the image plane.

\section{Extended Details of Homography-based Pseudo-label Mining}
The complete procedure of the proposed Homography-based Pseudo-label Mining(HPM) algorithm is summarized in Algorithm~\ref{hpm}. Several key components of this algorithm are explained below.

\noindent\textbf{Model Prediction}. Formally, the outputs of the teacher model $F$ for a  unlabeled image $I^{u}$ contains the 2D and 3D attributes for each predicted object:
\begin{equation}
 [Cls, BBox]_{2D}, [Points, Depth, Size, Ori]_{3D} = \boldsymbol{F}(\boldsymbol{I}^{u}).
\label{eq:outputs}
\end{equation}
where $Cls$ is the classification confidence, $BBox$ is the 2D bounding box of the object. $Points$ is the predicted projected points of the 3D bounding box in the image plane. In our main paper, we predict 10 points of the 3D box in total following MonoFlex\cite{monoflex}, which includes eight corner points and top and bottom center points. $Depth$ refers to the depth value of the bottom center point. $Size$ represents the 3D size and contains the length, width, and height of the object. $Ori$ represents the orientation of the object. \par
\noindent\textbf{2D-3D Transformation}. Homography-based Pseudo-label Mining involves the geometric transformation of the pseudo-labels between 2D and 3D space. Specifically, we take the four bottom corner points plus the bottom center point as the candidate points for each object to estimate the homography transformation. Specifically, the location of these candidate points in the image plane is directly obtained via Eq.\ref{eq:outputs}. To estimate the BEV coordinate of these points, the bottom center point is first transformed from the image plane to the camera coordinate system as:
\begin{equation}
 P_{center} = K^{-1}\cdot[zu, zv, z].
\label{eq:img_to_cam}
\end{equation}
Then, we apply the local transformation to the $P_c$ with the orientation and 3D size(length, width, and height) prediction in Eq.\ref{eq:outputs} to get the camera coordinates $P_{corner}$ of the candidate corner points. Such local transformation involves simply translation and rotation. Finally, the inversion of camera extrinsic matrix $[R|T]$ is further applied to obtain their BEV coordinates. 
\begin{equation}
 B_{\cdot} = [\mathbf{R} | \mathbf{T}]^{-1}\cdot P_{\cdot}
\label{eq:cam_to_bev}
\end{equation}
Where $P_{\cdot}$ refers to the camera coordinates of bottom corner points and bottom center point, and $B_{\cdot}$ denotes their corresponding BEV coordinates.\par
Note that the 3D BEV coordinate of these candidate points is not directly transformed by their 2D location in the image plane, instead, they are estimated based on the model's 3D attributes prediction such as depth, orientation, etc. Therefore, the homography matrix solved from these coordinates via DLT\cite{dlt} is not a trivial solution. 
\begin{algorithm*}[h]
	\caption{$\mathbf{DLT}$ indicates solving the homography matrix by Direct Linear Transform\cite{dlt}. $\mathbf{ImageCoord}_{\cdot}$ and $\mathbf{BEVCoord}_{\cdot}$ refers to the operation to obtain the coordinates of the point in the image plane and BEV plane. $\mathbf{ImageCoord}_{\cdot}$ relies on Eq.\ref{eq:outputs} and $\mathbf{BEVCoord}_{\cdot}$ relies on Eq.\ref{eq:img_to_cam},Eq.\ref{eq:cam_to_bev}.
 '$\mathbf{all}$' indicates all candidate points(5 points each object), '$\mathbf{bc}$' indicates the bottom center point. } 
	\label{hpm} 
 \SetKwInOut{Input}{\textbf{Input}}\SetKwInOut{Output}{\textbf{Output}}
         \Input{ \\
    $m^0 $: Initial set of pseudo-labels generated by uncertainty filtering; \\
    $\theta_h$: Localization error threshold; \\
    $ p = \{p_1, ..., p_N\}$: $N$ candidate predicted 3D bounding box}
    \Output{
        \\
        $m^{3D}$: The final pseudo-labels for 3D attributes \\}

\begin{algorithmic}[1]
\BlankLine

\FOR{$t=1,\cdots,t_{max}$}
    \STATE  $\tilde{C}_I^t = \mathbf{ImageCoord_{all}}( m^{t-1})$; // Image coordinates of all pseudo-labels' bottom corner and center points
    \STATE  $\tilde{C}_B^t = \mathbf{BEVCoord_{all}}( m^{t-1})$;  // BEV coordinates of all pseudo-labels' bottom corner and center points
    \STATE   $\tilde{M} \leftarrow \mathbf{DLT}(\tilde{C}_I^t, \tilde{C}_B^t)$;
    \STATE $m^{t} \leftarrow m^{t-1}$;
    \FOR{$j=1,\cdots,N$}
        \STATE $\hat{c}^b \leftarrow \tilde{M} [\mathbf{ImageCoord_{bc}}(p_j), 1]^T$;   // Obtain BEV coordinates of the bottom center points via homography transformation
        \STATE $\epsilon_j^t  \leftarrow ||\mathbf{BEVCoord_{bc}}(p_j) - \hat{c}^b ||_2$; // Compute loc error of bottom center point
        \IF{$\epsilon_j^t < \theta_h$}
          \STATE $m^{t} \leftarrow$ add $b_j$ into  $m^{t}$;
        \ENDIF
    \ENDFOR  
    \IF {$m^{t} = m^{t - 1}$}
        \STATE break;
    \ENDIF
\ENDFOR
\STATE  $m^{3D}  \leftarrow m^t$
\STATE return $m^{3D}$
 
	\end{algorithmic}

\end{algorithm*}
\par
\noindent\textbf{Feasibility of Flat Ground Assumption}. We check the feasibility of the flat ground assumption used in the homography-based pseudo-label mining algorithm. Actually, the KITTI dataset does exhibit some micro-unevenness in the ground, which also leads to minor localization errors in ground truth objects when utilizing the homography solved from the ground truth bounding box as shown in Tab.\ref{tab:localization_error}. However, these errors are minimal when compared to the localization errors in pseudo-labels caused by inaccurate depth. With this obvious gap between the ground truth bounding box and the pseudo-labels with inaccurate 3D attribute prediction, our method remains applicable to distinguish the reliable pseudo-labels. But it's worth noting that on a severely uneven road surface, where the homography constraint is substantially violated by the ground truth object, our approach may struggle to distinguish between reliable and unreliable pseudo-labels.

\section{Extended Implementation Details}
 Our experiments are conducted based on the MonoFlex\cite{monoflex} with the official code provided by the authors. For the \textbf{KITTI} dataset, we first pre-train the model on the labeled data for 140 epochs with a batch size of 8 following the default setting. After that, we copy the pre-trained model weight into the student and teacher models for end-to-end semi-supervised fine-tuning. For each iteration of semi-supervised fine-tuning, we randomly select 8 labeled images and 8 unlabeled images as the batched data and pad the images to the size of [1280, 384]. We utilize the AdamW optimizer with a learning rate of 3e-4, and weight decay of 1e-5, and fine-tune the model with semi-supervised learning for 20 epochs, in which the learning rate is decayed at the 10th and 15th epochs by a factor of 0.1, respectively. To demonstrate the generality of our approach, we also conduct experiments on the \textbf{nuScenes} dataset~\cite{nuscenes}, which is another large-scale autonomous driving dataset. Since~\cite{pgd,fcos3d} are the only M3OD works that provide the results on this benchmark, we choose to conduct experiments with these two base detectors. For the nuScenes dataset, we follow the default setting of FCOS3D\cite{fcos3d} and PGD\cite{pgd} implemented in MMDetection3D. We first pre-train the model on the labeled data for 12 epochs with a batch size of 16 and input size of [1600,900]. We utilize the SGD optimizer with a learning rate of 2e-3 and weight decay of 1e-4. For each iteration of semi-supervised fine-tuning, we randomly select 8 labeled images and 8 unlabeled images as the batched data and conduct fine-tuning for 5 epochs, in which the learning rate is decayed at the 2nd and 4th epochs by a factor of 0.1, respectively.
 For the experiments on Other M3OD Detectors, given that the pseudo-label mining algorithm based on homograph in DPL relies on key point prediction of the 3D bounding box. For the monocular 3D object detector \cite{monodetr,monodle,gupnet} without key point prediction, we add the key point prediction branch head to the original head. All experiments are conducted with 8$\times$ 32G NVIDIA Tesla V100 GPUs.

\section{More Experiment Results}
\begin{figure*}[ht]
    \centering
    \includegraphics[width=0.9\linewidth]{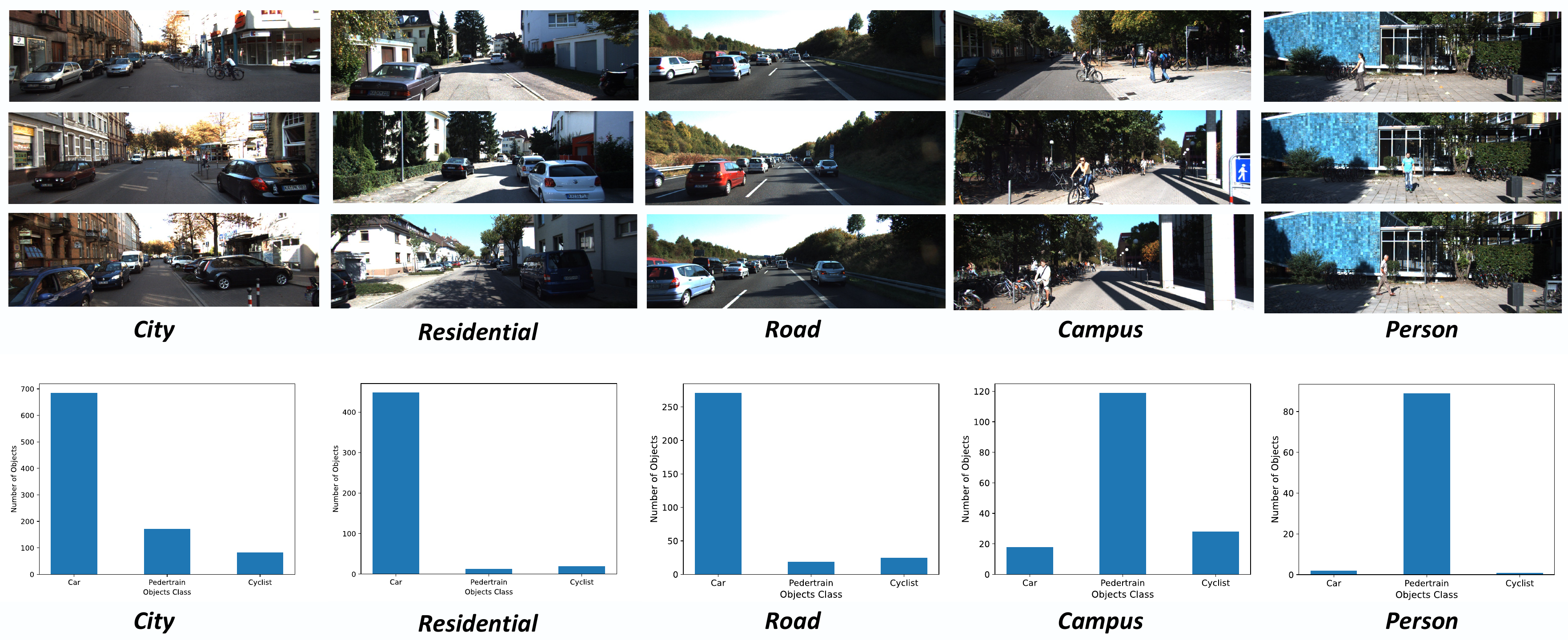}
    \caption{KITTI raw data spans five scene types. Bottom: Object class distribution across scenes. Accordingly, We divide the unlabeled data into two groups: (1) Car-dominated (city, residential, road), and (2) Person-dominated (campus, person).
    }
    \label{fig:kitti_diversity}
\end{figure*}

\noindent\textbf{Effect of Diversity of Unlabeled Data}. We analyze the impact on the SSM3OD performance of the distribution of the unlabeled data.
The KITTI raw data were collected from five diverse scenes: city, residential, road, campus, and person, as depicted in the right of Fig.\ref{fig:kitti_diversity}. Analyzing the object class distribution in each scene revealed significant differences. For example, in residential and road scenes, the car object dominates, with few pedestrians and cyclists. Conversely, campus and personal scenes mainly consist of pedestrians. To investigate the effect of unlabeled data diversity, the images were roughly divided into two groups: car-oriented, and person-oriented according to their object class distribution. We randomly chose 5K images for each group as the unlabeled data. We further constructed a more comprehensive unlabeled data set by combining the images randomly selected from both two groups, with 2.5K images selected for each group. The results are reported in Tab.\ref{diversity_performance}. It clearly shows that the diversity of these classes affects performance. Specifically, car-dominated unlabeled images boost the performance of the car category, while resulting in a slight performance decrease for the pedestrian and cyclist category. There is an opposite trend when training with person-dominated unlabeled images, in which the performance of pedestrian and cyclist categories are improved and no obvious performance gain for the car category is observed. The main reason behind these results is the confirmation bias caused by the class imbalance. By contrast, the unlabeled data combined images from both groups, containing rich objects, improve the performance of all three object categories. These results underscore the importance of unlabeled data diversity in semi-supervised learning for M3OD.

\noindent\textbf{Performance on Large Scale Dataset}.  The nuScenes Dataset is a large dataset for multi-view 3D object detection~(MVOD), and some M3OD methods~\cite{fcos3d,pgd} can be extended to achieve MVOD by conducting monocular detection in every single view and then fusing the multi-view detection results. To demonstrate the generality of our method, we conducted further experiments on this large-scale dataset based on FCOS3D and PGD with the official codes from MMDetection3D. Note MVC-MonoDet~\cite{mvc-monodet} is the only SSM3OD work that reports the results on the nuScenes dataset, but they only present some partial metrics. Our results, detailed in Tab.\ref{nuscense}, showcase substantial performance improvements through our proposed pseudo-labeling method. Specifically, we achieve gains of \textbf{3.1} in mAP and \textbf{2.2} in NDS for FCOS3D, \textbf{1.8} in mAP, and \textbf{1.2} in NDS for PGD. Our method also outperforms MVC-MonoDet in both mAP and mATE metrics.  These results verify the efficacy of our method, demonstrating its potential for extending to multi-view 3D object detection and generalization. Note that MVOD focuses on feature intersection between different views or temporal frames which is beyond the scope of this article. Therefore, we \textit{not aim to surpass the state-of-the-art methods~\cite{bevformer,bevdet} on this benchmark} 
, and instead just to show the generalization ability of our method.

\begin{table}[H]
\centering
\caption{The effect of the diversity of unlabeled data. For each group, we randomly select 5K images as the corresponding unlabeled data, respectively. We further construct a comprehensive unlabeled data set by combining both images randomly selected from Car-dominated and person-dominated groups, with 2.5K images randomly chosen for each group.}
\footnotesize
\label{diversity_performance}
\setlength\tabcolsep{2pt} 
\begin{tabular}{c|ccc|ccc|ccc}
\hline \multirow{3}{*}{ Unlabeled Data } & \multicolumn{9}{|c}{ Val, $A P_{3 D}|R_{40}$} \\
\cline{2-10}
 &\multicolumn{3}{|c}{Car}& \multicolumn{3}{|c}{ Pedestrian} &\multicolumn{3}{|c}{Cyclist}  \\
 \cline{2-10}
& Easy & Mod & Hard& Easy & Mod & Hard & Easy & Mod & Hard  \\
\hline
Sup-baseline &22.80& 17.51 & 14.90& 7.30& 5.53 & 4.24 & 4.67 & 2.23  &1.93\\
Car-dominated & \textbf{24.54}&18.44 & 15.68& 6.91& 5.43 & 4.51 & 3.52 &  2.04 &1.65\\
Person-dominated &22.76& 17.23 & 14.71&8.23& 6.66  & 5.02 & 4.87  &2.32 & 2.11\\
Combined &24.32& \textbf{18.56} & \textbf{16.12}&\textbf{8.45}& \textbf{6.72} & \textbf{5.07}& \textbf{5.35} & \textbf{2.89}  &\textbf{2.43}\\
\hline
\end{tabular}
\end{table}

\begin{table*}[h]
\centering

\caption{The performance comparison in the \textbf{nuScenes validation set}. Note that the MVC-MonoDet only provides the mAP and mATE metrics in this dataset.}
\setlength\tabcolsep{2pt} 

\begin{tabular}{c|c|c|c|c|c|c|c|c}
\hline
\label{nuscense}
\multirow{2}{*}{Methods}  & \multirow{2}{*}{Extra Data}  & mAP $\uparrow$ & mATE $\downarrow$ & mASE $\downarrow$ & mAOE $\downarrow$ &mAVE $\downarrow$ & mAAE $\downarrow$ & NDS $\uparrow$ \\
\cline{3-9}
& & (\%)&(m)&(1-iou)&(rad)&(m/s)&(1-acc)&\%\\
\hline MVC-MonoDet & Unlabeled & 0.349 & 0.640 & - & - & - & - & - \\
 FCOS3D & None & 0.321 & 0.754 & 0.260 & 0.486 & 1.332 & 0.157 & 0.394 \\
DPL$_{FCOS3D}$ & Unlabeled & 0.352 & 0.633 & \textbf{0.248} & 0.423 & 1.264 & \textbf{0.143} & 0.416 \\
 PGD & None & 0.358 & 0.667 & 0.264 & 0.434 & 1.276 & 0.176 & 0.425 \\
DPL$_{PGD }$&  Unlabeled & \textbf{0.376} & \textbf{0.577} & 0.250 & \textbf{0.412} & \textbf{1.258} & 0.161 & \textbf{0.437} \\
\hline
\end{tabular}
\end{table*}

\noindent\textbf{Ablation of Threshold}. We ablate the threshold of uncertainty threshold $\theta_u$, location error threshold $\theta_h$ in Tab.\ref{tab:ablation}. The best results achieve with $\theta_u = 0.10$ and $\theta_h = 2.0$.

\begin{figure}[h]
    \centering
    \begin{minipage}[c]{0.27\textwidth}
    \begin{table}[H]
        \centering
        \footnotesize
        \label{different_base_detectors}
        \setlength\tabcolsep{3pt} 
        \begin{tabular}{cc|c|c|c}
            \hline
            \multirow{2}{*}{$\theta_u$} & \multirow{2}{*}{$\theta_h$} & \multicolumn{3}{|c}{Val, $AP_{3D}|R_{40}$} \\
            \cline{3-5} & & Easy & Mod & Hard  \\
            \hline 
            0.10 & 1.0 & 25.68 & 19.39 & 17.06\\
            0.20 & 2.0 & 24.94 & 19.15 & 16.82 \\
            0.10 & 2.0 & \textbf{26.51} & \textbf{19.84} & \textbf{17.13}\\
            \hline
        \end{tabular}
        \caption{Ablation of the threshold $\theta_u$ and $\theta_h$.}
         \label{tab:ablation}
        \end{table}
    \end{minipage}
    \hfill
    \begin{minipage}[c]{0.2\textwidth}
        \centering
        \begin{table}[H]
        \footnotesize
        \begin{tabular}{c|c}
            \hline
            Labels & \makecell[c]{Loc Error(m)} \\
            \hline
            GTs  & 0.91 \\
            PLs   & 2.29 \\
            \hline
        \end{tabular}
        \caption{The average localization errors of pseudo labels(PL) and ground truth(GT) labels.}
        \label{tab:localization_error}
        \end{table}
    \end{minipage}
\end{figure}
\noindent\textbf{Performance on Pedestrian and Cyclist Categoriy}. We also report the detection performance on the Pedestrian and Cyclist categories on the KITTI test set in Tab.\ref{ped_and_cyclist}, where our method also provides a significant boost in detection performance for these categories with relatively few instances. 

\begin{table}
 \centering
\setlength\tabcolsep{3pt} 
\caption{Performance comparison on the \textbf{KITTI test} set of the Pedestrian and Cyclist category.}
\begin{tabular}{c|c|c|c|c|c|c}
\hline 
\label{ped_and_cyclist}
\multirow{3}{*}{ Methods } & \multicolumn{6}{|c}{ Test, $A P_{3 D}|R_{40}$} \\
\cline { 2 - 7 } & \multicolumn{3}{|c|}{ Pedestrian } & \multicolumn{3}{c}{ Cyclist } \\
\cline { 2 - 7 } & Easy & Mod & Hard & Easy & Mod & Hard \\
\hline M3D-RPN & 4.92 & 3.48 & 2.94 & 0.94 & 0.65 & 0.47 \\
MonoPair& 10.02 & 6.68 & 5.53 & 3.79 & 2.12 & 1.83 \\
MonoFlex † & 9.02 & 6.13 & 5.14 & 2.36& 1.44 &1.07 \\
\hline DPL$_{FLEX }$ & $\mathbf{11.66}$ & $\mathbf{7.52}$ & $\mathbf{6.16}$ & $\mathbf{8.41}$ & $\mathbf{4.51}$ & $\mathbf{3.59}$ \\
\hline
\end{tabular}
\end{table}

\begin{figure*}[ht]
    \centering    \includegraphics[width=0.7\linewidth]{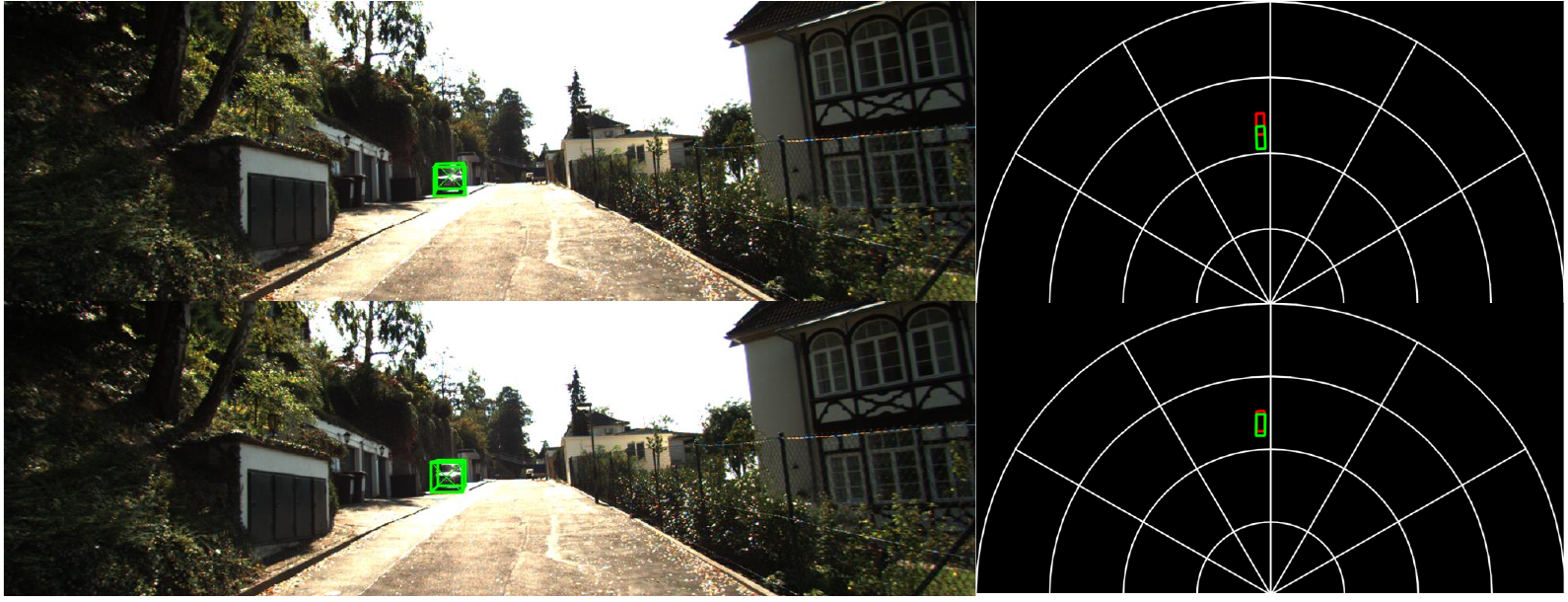}
\includegraphics[width=0.7\linewidth]{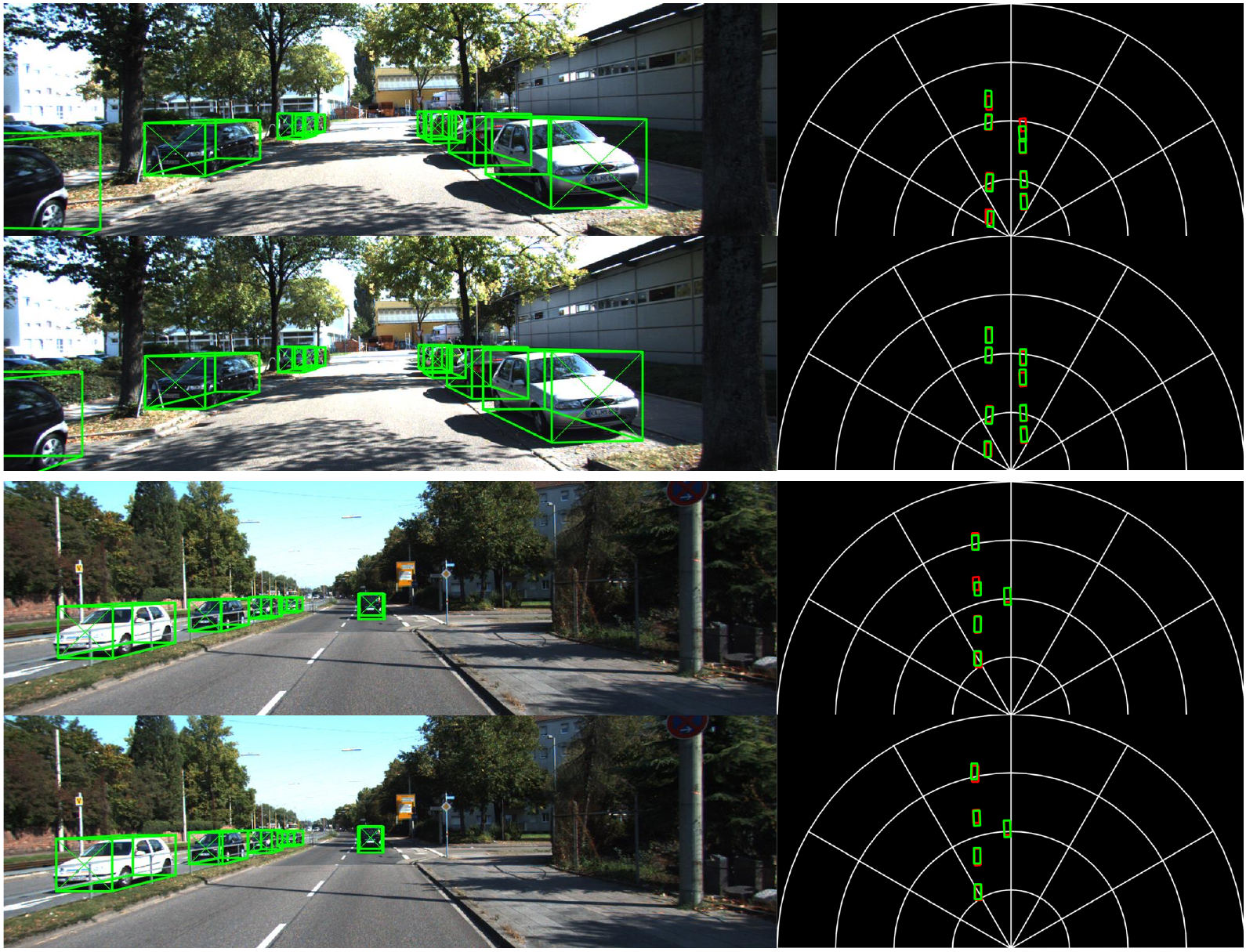}
\includegraphics[width=0.7\linewidth]{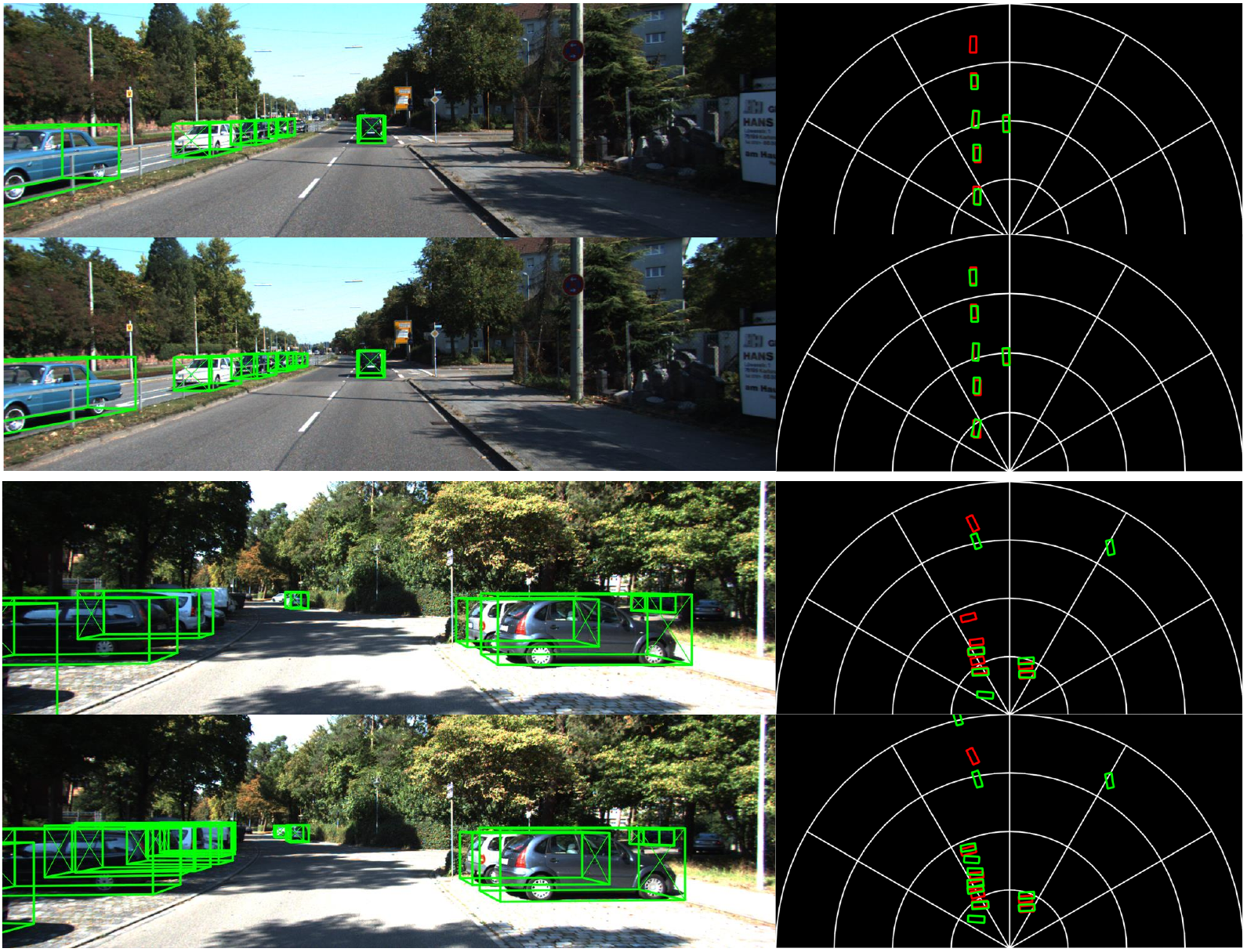}
\captionsetup{font=footnotesize}
    \caption{Visualization of the detection results on the KITTI validation set. For each image, the first and second rows are the detection results of the supervised baseline and our method, respectively. The box in \textcolor{red}{red} and \textcolor{green}{green} on the BEV plane are the ground truth box and detection bounding box, respectively.}
    \label{fig:kitti_vis}
\end{figure*}

\noindent\textbf{Detection Results Visualization}. We visualize the detection results of our method compared to the supervised baseline method on the KITTI validation set in Fig.\ref{fig:kitti_vis}. It clearly shows that our method not only detects objects more accurately, as observed in 1st, 2nd, and 3rd images but also exhibits higher prediction recall, as presented in 4th and 5th images. These results once again demonstrate the superiority of our method.

\section{Limitations}
Our method significantly improves the performance of monocular 3D detection methods with only image input. Compared with other 3D object detection methods, for example, LiDAR-based method, BEV-based method, etc, monocular 3D object exhibits a great advantage in the practical application.  With single-camera setups, it is more cost-effective and adaptable to numerous practical scenarios such as robotics, autonomous driving, and mobile augmented reality. However, due to the fundamental difficulty in estimating depth from a single RGB image, current performance still lags behind some methods using extra inputs (such as LiDAR). This limitation motivates us to explore the use of unlabeled data from other complementary sensor modalities such as LiDAR point clouds and stereo images. Actually, such kind of cross-modal learning has already validated its effectiveness in various tasks\cite{peng2021sparse}. For SSM3OD, these unlabeled data in other modalities contain more reliable object depth information, which can greatly ease the difficulty of accurately detecting 3D objects in real-world scenarios. We leave this exploration for our future work.


\end{document}